\documentclass{article}
\usepackage{booktabs} % 引入表格线条美化包
\usepackage{authblk}
\usepackage{amsmath}
\usepackage{amsfonts}
\usepackage{amssymb}
\usepackage{graphicx}
\usepackage{hyperref}
\usepackage{float}
\usepackage{multirow}
\usepackage{listings}
\usepackage[margin=1in]{geometry}
\hypersetup{
    colorlinks=true,
    linkcolor=blue,
    filecolor=magenta,
    urlcolor=cyan,
}
\usepackage[round, authoryear]{natbib} % 加载 natbib 宏包，设置引用样式为作者 - 年份且使用圆括号
\usepackage{url} % 支持处理 URL

\begin{document}

\title{1.4 Million Open-Source Distilled Reasoning Dataset to Empower Large Language Model Training}
\author{Han Zhao}
\author{Haotian Wang}
\author{Yiping Peng}
\author{Sitong Zhao}
\author{Xiaoyu Tian}
\author{Shuaiting Chen}
\author{Yunjie Ji}
\author{Xiangang Li}

\affil{
    \raisebox{-0.4em}{\includegraphics[height=1.5em]{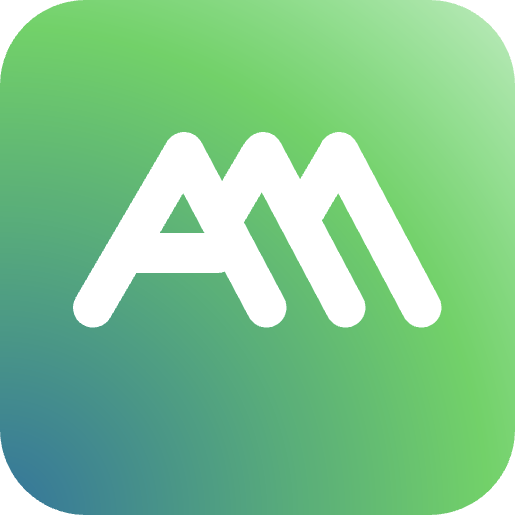}}
    \hspace{0.2em}a-m-team
}
\date{}

\maketitle

\begin{abstract}
\noindent The AM-DeepSeek-R1-Distilled is a large-scale dataset with thinking traces for general reasoning tasks, composed of high-quality and challenging reasoning problems. These problems are collected from a multitude of open-source datasets, subjected to semantic deduplication and meticulous cleaning to eliminate test set contamination. All responses within the dataset are distilled from reasoning models (predominantly DeepSeek-R1) and have undergone rigorous verification procedures. Mathematical problems are validated by checking against reference answers, code problems are verified using test cases, and other tasks are evaluated with the aid of a reward model. The AM-Distill-Qwen-32B model, which was trained through only simple Supervised Fine-Tuning (SFT) using this batch of data, outperformed the DeepSeek-R1-Distill-Qwen-32B model on four benchmarks: AIME2024, MATH-500, GPQA-Diamond, and LiveCodeBench. Additionally, the AM-Distill-Qwen-72B model surpassed the DeepSeek-R1-Distill-Llama-70B model on all benchmarks as well. We are releasing these 1.4 million problems and their corresponding responses to the research community with the objective of fostering the development of powerful reasoning-oriented Large Language Models (LLMs). The dataset was published in \href{https://huggingface.co/datasets/a-m-team/AM-DeepSeek-R1-Distilled-1.4M}{https://huggingface.co/datasets/a-m-team/AM-DeepSeek-R1-Distilled-1.4M}
\begin{figure}[h!]
    \centering
    \includegraphics[width=0.9\linewidth]{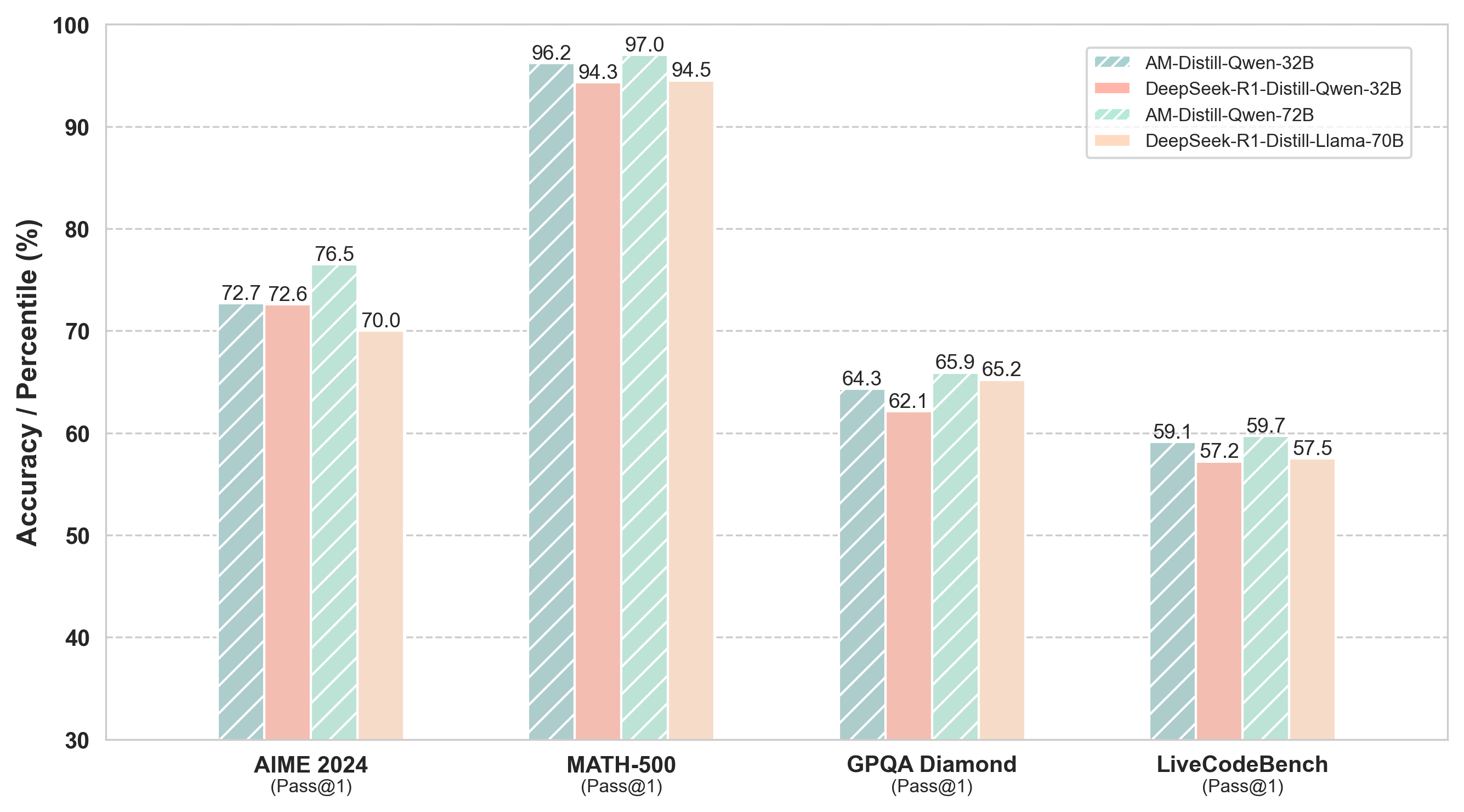}
    \caption{Overall performance of AM model}
    \label{fig:enter-label}
\end{figure}
\end{abstract}

\section{Introduction}
\setlength{\parskip}{0.5\baselineskip}
OpenAI’s o1 series models \citep{OpenAI2024} were the pioneers in introducing inference-time scaling by extending the length of the Chain-of-thought reasoning process \citep{wei2023chainofthoughtpromptingelicitsreasoning, snell2024scalingllmtesttimecompute, wu2025inferencescalinglawsempirical}. This approach has yielded remarkable improvements across various reasoning tasks, including mathematics, coding, and scientific reasoning \citep{lightman2023letsverifystepstep, hwang2024selfexploreenhancingmathematicalreasoning}.

Subsequently, the introduction of DeepSeek-R1 \citep{deepseekai2025deepseekr1incentivizingreasoningcapability} significantly propelled the open-source community forward, enabling deeper insights into inference-time scaling. DeepSeek also introduced the DeepSeek-R1-distilled series of models. These models solely utilized distilled data with reasoning chains for Supervised Fine-Tuning (SFT), yet they achieved outstanding results on diverse benchmarks. In the training pipeline of DeepSeek-R1, compared with DeepSeek-R1-Zero, 800,000 selected entries of data were used for SFT. This is a crucial factor contributing to DeepSeek-R1's superiority over DeepSeek-R1-Zero, thus demonstrating the necessity of high-quality SFT. SFT process, with carefully selected data, can effectively improve the performance of the model, as evidenced by the significant improvement of DeepSeek-R1 over its counterpart. This not only highlights the importance of data selection in SFT but also further validates the positive impact of well-executed SFT on enhancing a model's reasoning ability.

Building upon prior work, the open-source community has recently introduced numerous datasets that distilled reasoning models from DeepSeek-R1 \citep{openthoughts, xu2025kodcode}. However, the scale of these datasets is generally smaller than the 800,000 samples employed by DeepSeek in its distilled series of models. To date, few open-source initiatives have matched the performance achieved by the DeepSeek-R1-distilled series models based on the corresponding base models. Therefore, we have constructed the AM-DeepSeek-R1-Distilled dataset, which encompasses 1.4 million high-quality data entries with reasoning chains. Among these, 0.5 million data entries are entirely sourced from open-source datasets, and 0.9 million data entries are distilled by AM from DeepSeek-R1, as denoted by the ``am-0309'' in the response sources. The AM-DeepSeek-R1-Distilled dataset we developed exhibits significant advantages in terms of data scale, quality, and diversity. Through our meticulous data processing and stringent verification procedures, this dataset can offer robust support for the long COT training of large language models.

In terms of data collection, we comprehensively gathered diverse types of reasoning problems from numerous open-source datasets and implemented semantic deduplication and cleaning to guarantee the high quality and purity of the data \citep{li2023syntheticdatagenerationlarge, tirumala2023d4improvingllmpretraining}. Simultaneously, we conducted strict verification of all responses, including validating mathematical problems through answer checking, verifying code problems via test cases, and evaluating other tasks using a reward model, thereby ensuring the accuracy and reliability of the data.

Regarding data scale, the AM dataset, with its 1.4 million data entries, has significantly outperformed other recent open-source datasets. Among these entries, 500,000 are fully derived from open-source datasets. They span a wide range of knowledge domains and problem types. For the remaining 900,000, the instruction part is sourced from open-source datasets, and the response part is distilled by the AM team from DeepSeek-R1. These data have undergone processing in our data pipeline and possess high quality.

In terms of diversity, our dataset not only encompasses problems from common domains such as math, code, and science but also includes some cross-domain and comprehensive reasoning tasks. This can comprehensively exercise the reasoning ability and generalization ability of the models \citep{song2024scalingdatadiversityfinetuning}. Moreover, we meticulously processed the instruction part. We utilized a large language model to score all instructions in terms of difficulty and category and performed strict semantic deduplication according to these labels to ensure the high quality and diversity of the instructions \citep{xu2024magpiealignmentdatasynthesis}.

In addition, our dataset adopts a unified format, and each data entry is annotated in detail, including user-assistant interaction information, reasoning processes, final answers, reference answers, test cases, and other metadata. This standardized format renders the dataset easy to use and understand, facilitating researchers in conducting data processing and model training.

We believe that the release of the AM-DeepSeek-R1-Distilled dataset will offer crucial resource support for the research of reasoning-oriented large language models and is anticipated to drive further development and innovation in this field. We look forward to the research community leveraging this dataset to achieve more research breakthroughs and jointly promote the progress of AGI.

\section{Approach}
\begin{figure}[h!]
    \centering
    \includegraphics[width=1.0\linewidth]{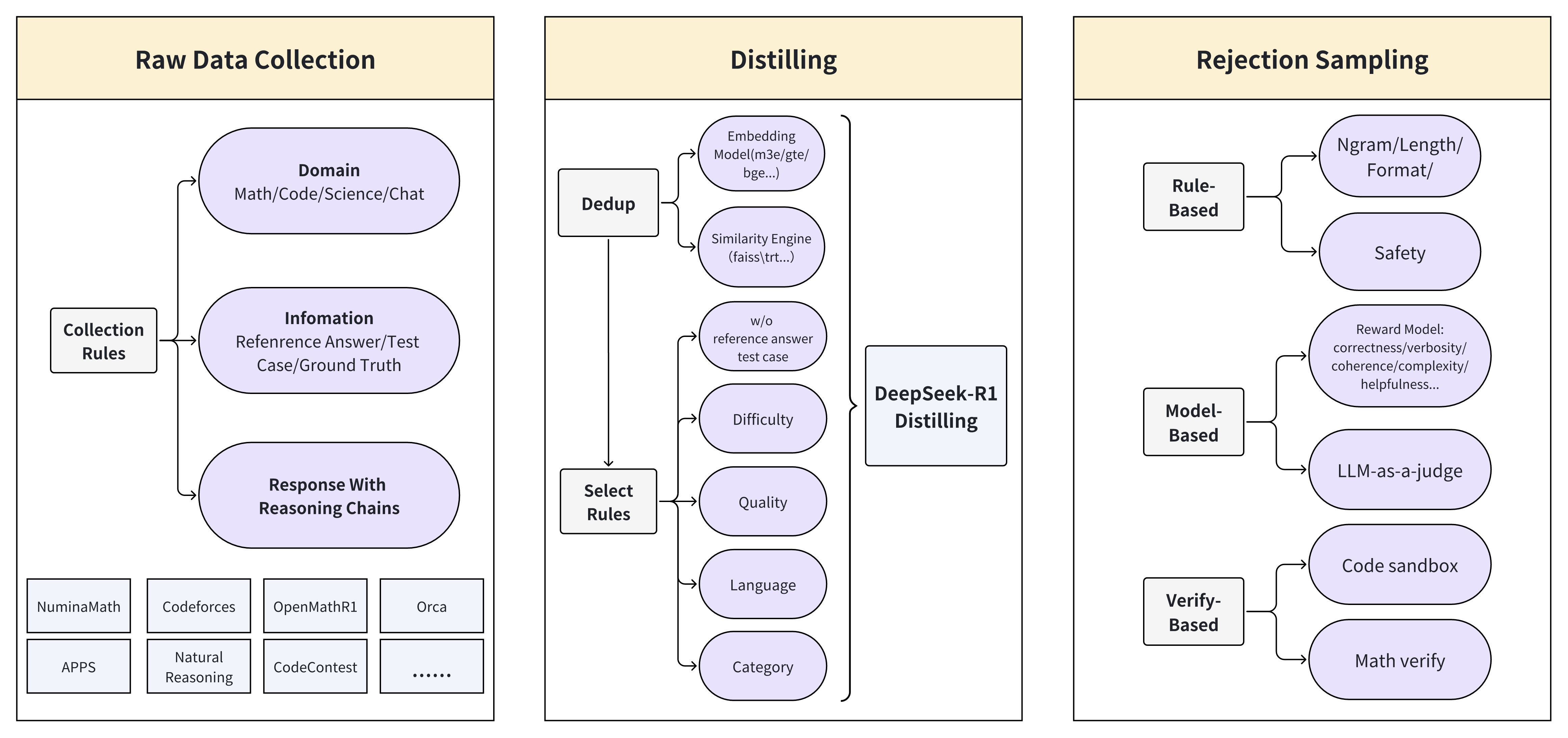}
        \caption{Construction process of data pipeline.}
    \label{fig:enter-label1}
\end{figure}

The core criteria for our data selection mainly include three aspects: diversity, complexity, and accuracy. We constructed our data pipeline around how to improve these three core indicators. As demonstrated in Figure~\ref{fig:enter-label1}, the entire pipeline can be divided into (1) Raw Data Collection, (2) Distilling, and (3) Rejection Sampling. The subsequent sections will elaborate on these components in detail.

\subsection{Raw Data}
\subsubsection{Data Sources}
We divided the data selection into four major categories: math, code, scienceQA, and general chat. We classified high-quality open-source datasets into these four categories. For Math, Code, and ScienceQA, we prioritized to select datasets with reference answers or test cases, such as NuminaMath \citep{numina_math_datasets}, MetaMathQA \citep{yu2023metamath}, natural\_reasoning \citep{yuan2025naturalreasoningreasoningwild28m}, OpenCoder \citep{Huang2024OpenCoderTO}, Omni-MATH \citep{gao2024omnimathuniversalolympiadlevel}, PRIME \citep{yuan2024implicitprm}, CodeIO \citep{li2025codeio}, MATH-lighteval \citep{hendrycksmath2021}. Additionally, we also selected some datasets with reasoning chains generated by DeepSeek-R1 from the open-source community, such as Openthoughts \citep{openthoughts}, OpenR1Math \citep{OpenR1Math}, KodCode \citep{xu2025kodcode}, Bespoke-Stratos-17k \citep{bespoke_stratos}, GeneralThought \citep{GeneralThought}, Dolphin-R1 \citep{dolphinr1}, data\_ablation\_full59K \citep{muennighoff2025s1simpletesttimescaling}, s1K \citep{muennighoff2025s1simpletesttimescaling}, LIMO \citep{ye2025limoreasoning}. Additionally, to enhance the model's chatting ability, we obtained chat data from general-data SFT datasets, such as InfinityInstruct \citep{InfinityInstruct2024}, Orca \citep{OpenOrca}. The distribution of reference answers and test cases can be found in Appendix \ref{a2}

\subsubsection{Categories}
The initial four categories alone were insufficient, especially for general chat data. Thus, we designed some more detailed categories, such as creative writing and instruction following. To facilitate data matching and enhance the diversity of the AM dataset, we used the Qwen2.5-7B-Instruct model \citep{qwen2.5} to label the data. The details of the categories can be found in the Appendix~\ref{a3} and Appendix~\ref{b2}.

\subsubsection{Difficulty}
For the training of long-cot models, more challenging data can effectively extend the length of the reasoning chains generated by the model and improve its reasoning ability. Thus, we used a large language model to score the difficulty of all instructions, subsequently screening the data and downsampling easy and medium difficulty examples. This ensures that the AM dataset emphasizes more challenging data while maintaining its diversity. The difficulty distribution of the data can be found in Appendix \ref{a4} and Appendix~\ref{b1}.

\subsubsection{Deduplication}
We performed strict semantic deduplication on the collected data. We calculated the embedding for each data entry and computed text similarity based on their embeddings to obtain the semantic similarity of different data entries. For data with high semantic similarity, we designed some priority strategies and ultimately retained only one representative entry. This process ensures dataset uniqueness and diversity of the dataset and prevents the negative impact of similar data during model training.

\subsection{Distilled Data}
We obtained responses to prompts via two ways: filtering existing responses and creating new responses. For prompts with existing responses, we retained the original response if it can pass reference-answer or test-case verification. For data without reasoning chains, we generated new responses using DeepSeek-R1.

\subsubsection{Ground Truth Verification}
For problems with available reference answers, we conducted verification through a combination of rule-based methods and a large language model. Initially, we applied math-verify \citep{MathVerify} to assess whether the response matched reference answers in terms of format and calculation results. Subsequently, we used Qwen2.5-7B-Instruct to further evaluate the correctness and consistency of these responses, the prompt could be found in Appendix~\ref{b3}. For code-related problems with test cases, we verified responses within a sandbox environment. We ultimately removed the data that did not pass the verification to ensure the accuracy and reliability of the dataset.

\subsubsection{Reward}
We used two methods, Decision-Tree-Reward-Llama-3.1-8B \citep{rlhflow2025decisiontree} as reward model and Qwen2.5-7B-Instruct for large language model scoring, to evaluate the answer\_content part of the model output. We set a certain score threshold based on the score distribution and removed the data with lower scores. The reward model evaluates responses across five dimensions: correctness, helpfulness, coherence, complexity, and verbosity to ensure the selected responses contribute to improving the overall quality of the dataset.

\subsubsection{Rule Verification}
We established verification rules, such as format template conformity and n-gram repetition checks. For format verification, we ensured that each response adhered strictly to the specified format, such as clearly indicating \textless think\textgreater reasoning process here\textless/think\textgreater \textless answer\textgreater final answer here\textless/answer\textgreater in the prompt. For n-gram repetition verification, we checked responses for excessive consecutive word repetition. Responses failing these rule-based verifications were excluded to guarantee dataset quality and consistency.

\enlargethispage{2\baselineskip}

\subsubsection{Labels}
We additionally annotated the data with supplementary information, such as length and language. For length annotation, we calculated the number of words or tokens per data entry, providing insights into the complexity and scale of the dataset. The length distribution of the data can be found in Appendix \ref{a1}. For language annotation, we primarily annotated entries as Chinese, English, or other languages. These labels facilitate effective data screening and analysis.

\section{Experiment}

\subsection{Evaluation}

\subsubsection{Benchmark}
We evaluated the reasoning ability of the model using LiveCodeBench \citep{jain2024livecodebench} (2024-08–2025-01), GPQA-Diamond \citep{rein2023gpqagraduatelevelgoogleproofqa}, AIME 2024 \citep{maa_aime_2024}, and MATH-500 \citep{lightman2023letsverifystepstep}. These benchmarks span multiple fields and difficulty levels, enabling a thorough assessment of the model's reasoning performance across diverse scenarios.

\subsubsection{Evaluation Methodology}
We set the maximum generation length to 32,768 tokens. For benchmarks requiring sampling, the temperature was uniformly set to 0.6, and the top-p value to 0.95. For AIME 2024 \citep{maa_aime_2024}, we generated 16 samples per query to estimate pass@1. For LiveCodeBench \citep{jain2024livecodebench}, MATH-500 \citep{lightman2023letsverifystepstep} and GPQA Diamond \citep{rein2023gpqagraduatelevelgoogleproofqa}, we generated 4 responses per query, also to estimate pass@1. The evaluation metric across these benchmarks was the globally averaged accuracy. 

\subsection{Main Result}
We performed SFT on Qwen2.5-32B producing a model named AM-Distill-Qwen-32B, the system prompt used is shown in Table~\ref{tab:system_prompt}. Compared with DeepSeek-R1-Distill-Qwen-32B, our models achieved significant improvements. Evaluation results are shown in Table~\ref{tab:model_performance}. Specifically, on AIME2024, the accuracy increased from 72.6\% to 72.7\%; on MATH-500, from 94.3\% to 96.2\%; on GPQA-Diamond, from 62.1\% to 64.3\%; and on LiveCodeBench, from 57.2\% to 59.1\%. Overall, the average accuracy improved from 71.6\% to 73.1\%.
\begin{table}[htbp]
  \centering
  \renewcommand{\arraystretch}{0.9} % 缩小行高，让内容更紧凑
  \begin{tabular}{p{1.0\textwidth}} 
    \toprule % 专业表格顶线
    \parbox{\linewidth}{
      You are a helpful assistant. To answer the user's question, you first think about the reasoning process and then provide the user with the answer. The reasoning process and answer are enclosed within \textless think\textgreater\ and \textless answer\textgreater\ tags, respectively, i.e., \textless think\textgreater\ reasoning process here \textless /think\textgreater\ \textless answer\textgreater\ answer here \textless /answer\textgreater.
    }\\
    \bottomrule % 专业表格底线
  \end{tabular}
  \caption{System prompt in training process.}
  \label{tab:system_prompt}
\end{table}

We further performed training based on the Qwen2.5-72B model to obtain AM-Distill-Qwen-72B. Compared with DeepSeek-R1-Distill-Llama-70B, our 72B model achieved notable improvements across all evaluation benchmarks. Specifically, accuracy on AIME2024 significantly increased from 70.0\% to 76.5\%; MATH-500 improved from 94.5\% to 97.0\%; GPQA-Diamond rose from 65.2\% to 65.9\%; and LiveCodeBench increased from 57.5\% to 59.7\%.

Experimental results demonstrate that models trained on our constructed AM-DeepSeek-R1-Distilled-1.4M dataset exhibit substantial enhancements in reasoning ability.

\begin{table}[htbp]
  \centering
  \begin{tabular}{l c c c c c}
        \hline
        Model & AIME2024 & MATH-500 & GPQA-Diamond & LiveCodeBench & Average \\
        \hline
        DeepSeek-R1-Distill-Qwen-32B & 72.6 & 94.3 & 62.1 & 57.2 & 71.6 \\
        AM-Distill-Qwen-32B & \textbf{72.7} & \textbf{96.2} & \textbf{64.3} & \textbf{59.1} & \textbf{73.1} \\
        \hline
        DeepSeek-R1-Distill-Llama-70B & 70.0 & 94.5 & 65.2 & 57.5 & 71.8 \\
        AM-Distill-Qwen-72B & \textbf{76.5} & \textbf{97.0} & \textbf{65.9} & \textbf{59.7} & \textbf{74.8} \\
        \hline
    \end{tabular}
    \caption{Model performance.}
    \label{tab:model_performance}
\end{table}

\section{Limitation}
Since the responses in this dataset are generated by large language models and have not been rigorously verified, there are still deficiencies in terms of factual accuracy and other aspects. When using this dataset, it is necessary to conduct a careful examination. This dataset is mainly used to enhance the reasoning capabilities of large language models (LLMs). We have not carried out a thorough filtering of the harmful instructions or responses within it. We require developers to use only the open-sourced code, data, model, and any other artifacts generated through this project for research purposes. Commercial use and other potential harmful use cases are not permitted. In addition, due to the nested relationships among some data sources, there may be issues with the inaccuracy of the data sources.

\section{Conclusion}
In this study, we have constructed and released an AM-DeepSeek-R1-Distilled dataset, a large-scale general reasoning task dataset with 1.4 million data entries and rich thinking traces. It was created through meticulous selection, semantic deduplication, and strict cleaning of a large number of open-source datasets.

Furthermore, the AM-Distill-Qwen-32B model, developed by performing SFT on Qwen2.5-32B with the utilization of our constructed dataset, has exhibited remarkable performance enhancements. This compellingly demonstrates that our dataset serves as a significant asset in training the reasoning capabilities of the model. We are optimistic that our endeavors will play a substantial and catalytic role in the research related to reasoning-oriented Large Language Models, propelling forward the development in this field.

\bibliographystyle{plainnat}
\bibliography{references}

\begin{thebibliography}{35}
\providecommand{\natexlab}[1]{#1}
\providecommand{\url}[1]{\texttt{#1}}
\expandafter\ifx\csname urlstyle\endcsname\relax
  \providecommand{\doi}[1]{doi: #1}\else
  \providecommand{\doi}{doi: \begingroup \urlstyle{rm}\Url}\fi

\bibitem[BAAI(2024)]{InfinityInstruct2024}
BAAI.
\newblock Infinity instruct.
\newblock 2024.
\newblock URL \url{https://huggingface.co/datasets/BAAI/Infinity-Instruct}.

\bibitem[Bespoke(2025)]{bespoke_stratos}
Bespoke.
\newblock Bespoke-stratos: The unreasonable effectiveness of reasoning distillation.
\newblock https://www.bespokelabs.ai/blog/bespoke-stratos-the-unreasonable-effectiveness-of-reasoning-distillation, 2025.
\newblock Accessed: 2025-01-22.

\bibitem[cognitivecomputations(2025)]{dolphinr1}
cognitivecomputations.
\newblock dolphin-r1.
\newblock \url{https://huggingface.co/datasets/cognitivecomputations/dolphin-r1}, 2025.

\bibitem[DeepSeek-AI et~al.(2025)DeepSeek-AI, Guo, Yang, Zhang, Song, Zhang, Xu, Zhu, Ma, Wang, Bi, Zhang, Yu, Wu, Wu, Gou, Shao, Li, Gao, Liu, Xue, Wang, Wu, Feng, Lu, Zhao, Deng, Zhang, Ruan, Dai, Chen, Ji, Li, Lin, Dai, Luo, Hao, Chen, Li, Zhang, Bao, Xu, Wang, Ding, Xin, Gao, Qu, Li, Guo, Li, Wang, Chen, Yuan, Qiu, Li, Cai, Ni, Liang, Chen, Dong, Hu, Gao, Guan, Huang, Yu, Wang, Zhang, Zhao, Wang, Zhang, Xu, Xia, Zhang, Zhang, Tang, Li, Wang, Li, Tian, Huang, Zhang, Wang, Chen, Du, Ge, Zhang, Pan, Wang, Chen, Jin, Chen, Lu, Zhou, Chen, Ye, Wang, Yu, Zhou, Pan, Li, Zhou, Wu, Ye, Yun, Pei, Sun, Wang, Zeng, Zhao, Liu, Liang, Gao, Yu, Zhang, Xiao, An, Liu, Wang, Chen, Nie, Cheng, Liu, Xie, Liu, Yang, Li, Su, Lin, Li, Jin, Shen, Chen, Sun, Wang, Song, Zhou, Wang, Shan, Li, Wang, Wei, Zhang, Xu, Li, Zhao, Sun, Wang, Yu, Zhang, Shi, Xiong, He, Piao, Wang, Tan, Ma, Liu, Guo, Ou, Wang, Gong, Zou, He, Xiong, Luo, You, Liu, Zhou, Zhu, Xu, Huang, Li, Zheng, Zhu, Ma, Tang, Zha, Yan, Ren, Ren, Sha, Fu, Xu, Xie, Zhang,
  Hao, Ma, Yan, Wu, Gu, Zhu, Liu, Li, Xie, Song, Pan, Huang, Xu, Zhang, and Zhang]{deepseekai2025deepseekr1incentivizingreasoningcapability}
DeepSeek-AI, Daya Guo, Dejian Yang, Haowei Zhang, Junxiao Song, Ruoyu Zhang, Runxin Xu, Qihao Zhu, Shirong Ma, Peiyi Wang, Xiao Bi, Xiaokang Zhang, Xingkai Yu, Yu~Wu, Z.~F. Wu, Zhibin Gou, Zhihong Shao, Zhuoshu Li, Ziyi Gao, Aixin Liu, Bing Xue, Bingxuan Wang, Bochao Wu, Bei Feng, Chengda Lu, Chenggang Zhao, Chengqi Deng, Chenyu Zhang, Chong Ruan, Damai Dai, Deli Chen, Dongjie Ji, Erhang Li, Fangyun Lin, Fucong Dai, Fuli Luo, Guangbo Hao, Guanting Chen, Guowei Li, H.~Zhang, Han Bao, Hanwei Xu, Haocheng Wang, Honghui Ding, Huajian Xin, Huazuo Gao, Hui Qu, Hui Li, Jianzhong Guo, Jiashi Li, Jiawei Wang, Jingchang Chen, Jingyang Yuan, Junjie Qiu, Junlong Li, J.~L. Cai, Jiaqi Ni, Jian Liang, Jin Chen, Kai Dong, Kai Hu, Kaige Gao, Kang Guan, Kexin Huang, Kuai Yu, Lean Wang, Lecong Zhang, Liang Zhao, Litong Wang, Liyue Zhang, Lei Xu, Leyi Xia, Mingchuan Zhang, Minghua Zhang, Minghui Tang, Meng Li, Miaojun Wang, Mingming Li, Ning Tian, Panpan Huang, Peng Zhang, Qiancheng Wang, Qinyu Chen, Qiushi Du, Ruiqi Ge, Ruisong
  Zhang, Ruizhe Pan, Runji Wang, R.~J. Chen, R.~L. Jin, Ruyi Chen, Shanghao Lu, Shangyan Zhou, Shanhuang Chen, Shengfeng Ye, Shiyu Wang, Shuiping Yu, Shunfeng Zhou, Shuting Pan, S.~S. Li, Shuang Zhou, Shaoqing Wu, Shengfeng Ye, Tao Yun, Tian Pei, Tianyu Sun, T.~Wang, Wangding Zeng, Wanjia Zhao, Wen Liu, Wenfeng Liang, Wenjun Gao, Wenqin Yu, Wentao Zhang, W.~L. Xiao, Wei An, Xiaodong Liu, Xiaohan Wang, Xiaokang Chen, Xiaotao Nie, Xin Cheng, Xin Liu, Xin Xie, Xingchao Liu, Xinyu Yang, Xinyuan Li, Xuecheng Su, Xuheng Lin, X.~Q. Li, Xiangyue Jin, Xiaojin Shen, Xiaosha Chen, Xiaowen Sun, Xiaoxiang Wang, Xinnan Song, Xinyi Zhou, Xianzu Wang, Xinxia Shan, Y.~K. Li, Y.~Q. Wang, Y.~X. Wei, Yang Zhang, Yanhong Xu, Yao Li, Yao Zhao, Yaofeng Sun, Yaohui Wang, Yi~Yu, Yichao Zhang, Yifan Shi, Yiliang Xiong, Ying He, Yishi Piao, Yisong Wang, Yixuan Tan, Yiyang Ma, Yiyuan Liu, Yongqiang Guo, Yuan Ou, Yuduan Wang, Yue Gong, Yuheng Zou, Yujia He, Yunfan Xiong, Yuxiang Luo, Yuxiang You, Yuxuan Liu, Yuyang Zhou, Y.~X. Zhu,
  Yanhong Xu, Yanping Huang, Yaohui Li, Yi~Zheng, Yuchen Zhu, Yunxian Ma, Ying Tang, Yukun Zha, Yuting Yan, Z.~Z. Ren, Zehui Ren, Zhangli Sha, Zhe Fu, Zhean Xu, Zhenda Xie, Zhengyan Zhang, Zhewen Hao, Zhicheng Ma, Zhigang Yan, Zhiyu Wu, Zihui Gu, Zijia Zhu, Zijun Liu, Zilin Li, Ziwei Xie, Ziyang Song, Zizheng Pan, Zhen Huang, Zhipeng Xu, Zhongyu Zhang, and Zhen Zhang.
\newblock Deepseek-r1: Incentivizing reasoning capability in llms via reinforcement learning, 2025.
\newblock URL \url{https://arxiv.org/abs/2501.12948}.

\bibitem[Gao et~al.(2024)Gao, Song, Yang, Cai, Miao, Dong, Li, Ma, Chen, Xu, Tang, Wang, Zan, Quan, Zhang, Sha, Zhang, Ren, Liu, and Chang]{gao2024omnimathuniversalolympiadlevel}
Bofei Gao, Feifan Song, Zhe Yang, Zefan Cai, Yibo Miao, Qingxiu Dong, Lei Li, Chenghao Ma, Liang Chen, Runxin Xu, Zhengyang Tang, Benyou Wang, Daoguang Zan, Shanghaoran Quan, Ge~Zhang, Lei Sha, Yichang Zhang, Xuancheng Ren, Tianyu Liu, and Baobao Chang.
\newblock Omni-math: A universal olympiad level mathematic benchmark for large language models, 2024.
\newblock URL \url{https://arxiv.org/abs/2410.07985}.

\bibitem[Hendrycks et~al.(2021)Hendrycks, Burns, Kadavath, Arora, Basart, Tang, Song, and Steinhardt]{hendrycksmath2021}
Dan Hendrycks, Collin Burns, Saurav Kadavath, Akul Arora, Steven Basart, Eric Tang, Dawn Song, and Jacob Steinhardt.
\newblock Measuring mathematical problem solving with the math dataset.
\newblock \emph{arXiv preprint arXiv:2103.03874}, 2021.

\bibitem[Huang et~al.(2024)Huang, Cheng, Liu, Hao, Song, Xu, Yang, Liu, Zhang, Chai, Yuan, Zhang, Fu, Liu, Zhang, Wang, Qi, Xu, and Chu]{Huang2024OpenCoderTO}
Siming Huang, Tianhao Cheng, Jason~Klein Liu, Jiaran Hao, Liuyihan Song, Yang Xu, J.~Yang, J.~H. Liu, Chenchen Zhang, Linzheng Chai, Ruifeng Yuan, Zhaoxiang Zhang, Jie Fu, Qian Liu, Ge~Zhang, Zili Wang, Yuan Qi, Yinghui Xu, and Wei Chu.
\newblock Opencoder: The open cookbook for top-tier code large language models.
\newblock 2024.
\newblock URL \url{https://arxiv.org/pdf/2411.04905}.

\bibitem[Hwang et~al.(2024)Hwang, Kim, Kim, Ye, and Seo]{hwang2024selfexploreenhancingmathematicalreasoning}
Hyeonbin Hwang, Doyoung Kim, Seungone Kim, Seonghyeon Ye, and Minjoon Seo.
\newblock Self-explore: Enhancing mathematical reasoning in language models with fine-grained rewards, 2024.
\newblock URL \url{https://arxiv.org/abs/2404.10346}.

\bibitem[Jain et~al.(2024)Jain, Han, Gu, Li, Yan, Zhang, Wang, Solar-Lezama, Sen, and Stoica]{jain2024livecodebench}
Naman Jain, King Han, Alex Gu, Wen-Ding Li, Fanjia Yan, Tianjun Zhang, Sida Wang, Armando Solar-Lezama, Koushik Sen, and Ion Stoica.
\newblock Livecodebench: Holistic and contamination free evaluation of large language models for code.
\newblock \emph{arXiv preprint arXiv:2403.07974}, 2024.

\bibitem[Kydlíček and Gandenberger(2025)]{MathVerify}
Hynek Kydlíček and Greg Gandenberger.
\newblock Math-verify, 2025.
\newblock URL \url{https://github.com/huggingface/math-verify}.

\bibitem[LI et~al.(2024)LI, Beeching, Tunstall, Lipkin, Soletskyi, Huang, Rasul, Yu, Jiang, Shen, Qin, Dong, Zhou, Fleureau, Lample, and Polu]{numina_math_datasets}
Jia LI, Edward Beeching, Lewis Tunstall, Ben Lipkin, Roman Soletskyi, Shengyi~Costa Huang, Kashif Rasul, Longhui Yu, Albert Jiang, Ziju Shen, Zihan Qin, Bin Dong, Li~Zhou, Yann Fleureau, Guillaume Lample, and Stanislas Polu.
\newblock Numinamath.
\newblock \url{https://huggingface.co/AI-MO/NuminaMath-CoT}, 2024.

\bibitem[Li et~al.(2025)Li, Guo, Yang, Xu, Wu, and He]{li2025codeio}
Junlong Li, Daya Guo, Dejian Yang, Runxin Xu, Yu~Wu, and Junxian He.
\newblock Codei/o: Condensing reasoning patterns via code input-output prediction.
\newblock \emph{arXiv preprint arXiv:2502.07316}, 2025.

\bibitem[Li et~al.(2023)Li, Zhu, Lu, and Yin]{li2023syntheticdatagenerationlarge}
Zhuoyan Li, Hangxiao Zhu, Zhuoran Lu, and Ming Yin.
\newblock Synthetic data generation with large language models for text classification: Potential and limitations, 2023.
\newblock URL \url{https://arxiv.org/abs/2310.07849}.

\bibitem[Lian et~al.(2023)Lian, Goodson, Pentland, Cook, Vong, and "Teknium"]{OpenOrca}
Wing Lian, Bleys Goodson, Eugene Pentland, Austin Cook, Chanvichet Vong, and "Teknium".
\newblock Openorca: An open dataset of gpt augmented flan reasoning traces.
\newblock \url{https://https://huggingface.co/datasets/Open-Orca/OpenOrca}, 2023.

\bibitem[Lightman et~al.(2023)Lightman, Kosaraju, Burda, Edwards, Baker, Lee, Leike, Schulman, Sutskever, and Cobbe]{lightman2023letsverifystepstep}
Hunter Lightman, Vineet Kosaraju, Yura Burda, Harri Edwards, Bowen Baker, Teddy Lee, Jan Leike, John Schulman, Ilya Sutskever, and Karl Cobbe.
\newblock Let's verify step by step, 2023.
\newblock URL \url{https://arxiv.org/abs/2305.20050}.

\bibitem[MAA(2024)]{maa_aime_2024}
MAA.
\newblock American invitational mathematics examination - aime.
\newblock \url{https://maa.org/math-competitions/american-invitational-mathematics-examination-aime}, feb 2024.
\newblock Accessed in February 2024, from American Invitational Mathematics Examination - AIME 2024.

\bibitem[Muennighoff et~al.(2025)Muennighoff, Yang, Shi, Li, Fei-Fei, Hajishirzi, Zettlemoyer, Liang, Candès, and Hashimoto]{muennighoff2025s1simpletesttimescaling}
Niklas Muennighoff, Zitong Yang, Weijia Shi, Xiang~Lisa Li, Li~Fei-Fei, Hannaneh Hajishirzi, Luke Zettlemoyer, Percy Liang, Emmanuel Candès, and Tatsunori Hashimoto.
\newblock s1: Simple test-time scaling, 2025.
\newblock URL \url{https://arxiv.org/abs/2501.19393}.

\bibitem[Open-R1(2025)]{OpenR1Math}
Open-R1.
\newblock Openr1-math-220k.
\newblock \url{https://huggingface.co/datasets/open-r1/OpenR1-Math-220k}, 2025.

\bibitem[OpenAI(2024)]{OpenAI2024}
OpenAI.
\newblock Learning to reason with llms, 2024.
\newblock URL \url{https://openai.com/index/learning-to-reason-with-llms/}.

\bibitem[OpenThoughts(2025)]{openthoughts}
Team OpenThoughts.
\newblock {Open Thoughts}.
\newblock https://open-thoughts.ai, January 2025.

\bibitem[Qwen(2024)]{qwen2.5}
Qwen.
\newblock Team qwen2.5: A party of foundation models, September 2024.
\newblock URL \url{https://qwenlm.github.io/blog/qwen2.5/}.

\bibitem[Reasoning(2025)]{GeneralThought}
General Reasoning.
\newblock Generalthought-feb25.
\newblock \url{https://huggingface.co/datasets/GeneralReasoning/GeneralThought-Feb25}, 02 2025.

\bibitem[Rein et~al.(2023)Rein, Hou, Stickland, Petty, Pang, Dirani, Michael, and Bowman]{rein2023gpqagraduatelevelgoogleproofqa}
David Rein, Betty~Li Hou, Asa~Cooper Stickland, Jackson Petty, Richard~Yuanzhe Pang, Julien Dirani, Julian Michael, and Samuel~R. Bowman.
\newblock Gpqa: A graduate-level google-proof q\&a benchmark, 2023.
\newblock URL \url{https://arxiv.org/abs/2311.12022}.

\bibitem[Rlhflow(2025)]{rlhflow2025decisiontree}
Rlhflow.
\newblock Decision tree reward model.
\newblock \url{https://rlhflow.github.io/posts/2025-01-22-decision-tree-reward-model/}, 2025.
\newblock Accessed: 2025-03-11.

\bibitem[Snell et~al.(2024)Snell, Lee, Xu, and Kumar]{snell2024scalingllmtesttimecompute}
Charlie Snell, Jaehoon Lee, Kelvin Xu, and Aviral Kumar.
\newblock Scaling llm test-time compute optimally can be more effective than scaling model parameters, 2024.
\newblock URL \url{https://arxiv.org/abs/2408.03314}.

\bibitem[Song et~al.(2024)Song, Yu, Lang, Yu, Huang, Wang, and Li]{song2024scalingdatadiversityfinetuning}
Feifan Song, Bowen Yu, Hao Lang, Haiyang Yu, Fei Huang, Houfeng Wang, and Yongbin Li.
\newblock Scaling data diversity for fine-tuning language models in human alignment, 2024.
\newblock URL \url{https://arxiv.org/abs/2403.11124}.

\bibitem[Tirumala et~al.(2023)Tirumala, Simig, Aghajanyan, and Morcos]{tirumala2023d4improvingllmpretraining}
Kushal Tirumala, Daniel Simig, Armen Aghajanyan, and Ari~S. Morcos.
\newblock D4: Improving llm pretraining via document de-duplication and diversification, 2023.
\newblock URL \url{https://arxiv.org/abs/2308.12284}.

\bibitem[Wei et~al.(2023)Wei, Wang, Schuurmans, Bosma, Ichter, Xia, Chi, Le, and Zhou]{wei2023chainofthoughtpromptingelicitsreasoning}
Jason Wei, Xuezhi Wang, Dale Schuurmans, Maarten Bosma, Brian Ichter, Fei Xia, Ed~Chi, Quoc Le, and Denny Zhou.
\newblock Chain-of-thought prompting elicits reasoning in large language models, 2023.
\newblock URL \url{https://arxiv.org/abs/2201.11903}.

\bibitem[Wu et~al.(2025)Wu, Sun, Li, Welleck, and Yang]{wu2025inferencescalinglawsempirical}
Yangzhen Wu, Zhiqing Sun, Shanda Li, Sean Welleck, and Yiming Yang.
\newblock Inference scaling laws: An empirical analysis of compute-optimal inference for problem-solving with language models, 2025.
\newblock URL \url{https://arxiv.org/abs/2408.00724}.

\bibitem[Xu et~al.(2024)Xu, Jiang, Niu, Deng, Poovendran, Choi, and Lin]{xu2024magpiealignmentdatasynthesis}
Zhangchen Xu, Fengqing Jiang, Luyao Niu, Yuntian Deng, Radha Poovendran, Yejin Choi, and Bill~Yuchen Lin.
\newblock Magpie: Alignment data synthesis from scratch by prompting aligned llms with nothing, 2024.
\newblock URL \url{https://arxiv.org/abs/2406.08464}.

\bibitem[Xu et~al.(2025)Xu, Liu, Yin, Zhou, and Poovendran]{xu2025kodcode}
Zhangchen Xu, Yang Liu, Yueqin Yin, Mingyuan Zhou, and Radha Poovendran.
\newblock Kodcode: A diverse, challenging, and verifiable synthetic dataset for coding.
\newblock 2025.
\newblock URL \url{https://arxiv.org/abs/2503.02951}.

\bibitem[Ye et~al.(2025)Ye, Huang, Xiao, Chern, Xia, and Liu]{ye2025limoreasoning}
Yixin Ye, Zhen Huang, Yang Xiao, Ethan Chern, Shijie Xia, and Pengfei Liu.
\newblock Limo: Less is more for reasoning, 2025.
\newblock URL \url{https://arxiv.org/abs/2502.03387}.

\bibitem[Yu et~al.(2023)Yu, Jiang, Shi, Yu, Liu, Zhang, Kwok, Li, Weller, and Liu]{yu2023metamath}
Longhui Yu, Weisen Jiang, Han Shi, Jincheng Yu, Zhengying Liu, Yu~Zhang, James~T Kwok, Zhenguo Li, Adrian Weller, and Weiyang Liu.
\newblock Metamath: Bootstrap your own mathematical questions for large language models.
\newblock \emph{arXiv preprint arXiv:2309.12284}, 2023.

\bibitem[Yuan et~al.(2024)Yuan, Li, Chen, Cui, Ding, Zhang, Zhou, Liu, and Peng]{yuan2024implicitprm}
Lifan Yuan, Wendi Li, Huayu Chen, Ganqu Cui, Ning Ding, Kaiyan Zhang, Bowen Zhou, Zhiyuan Liu, and Hao Peng.
\newblock Free process rewards without process labels.
\newblock \emph{arXiv preprint arXiv:2412.01981}, 2024.

\bibitem[Yuan et~al.(2025)Yuan, Yu, Jiang, Padthe, Li, Wang, Kulikov, Cho, Tian, Weston, and Li]{yuan2025naturalreasoningreasoningwild28m}
Weizhe Yuan, Jane Yu, Song Jiang, Karthik Padthe, Yang Li, Dong Wang, Ilia Kulikov, Kyunghyun Cho, Yuandong Tian, Jason~E Weston, and Xian Li.
\newblock Naturalreasoning: Reasoning in the wild with 2.8m challenging questions, 2025.
\newblock URL \url{https://arxiv.org/abs/2502.13124}.

\end{thebibliography}

\clearpage
\appendix

\section{Data Analysis}

\subsection{Length Distribution}
\label{a1}
\begin{figure}[ht]
    \centering
    \includegraphics[width=0.7\linewidth]{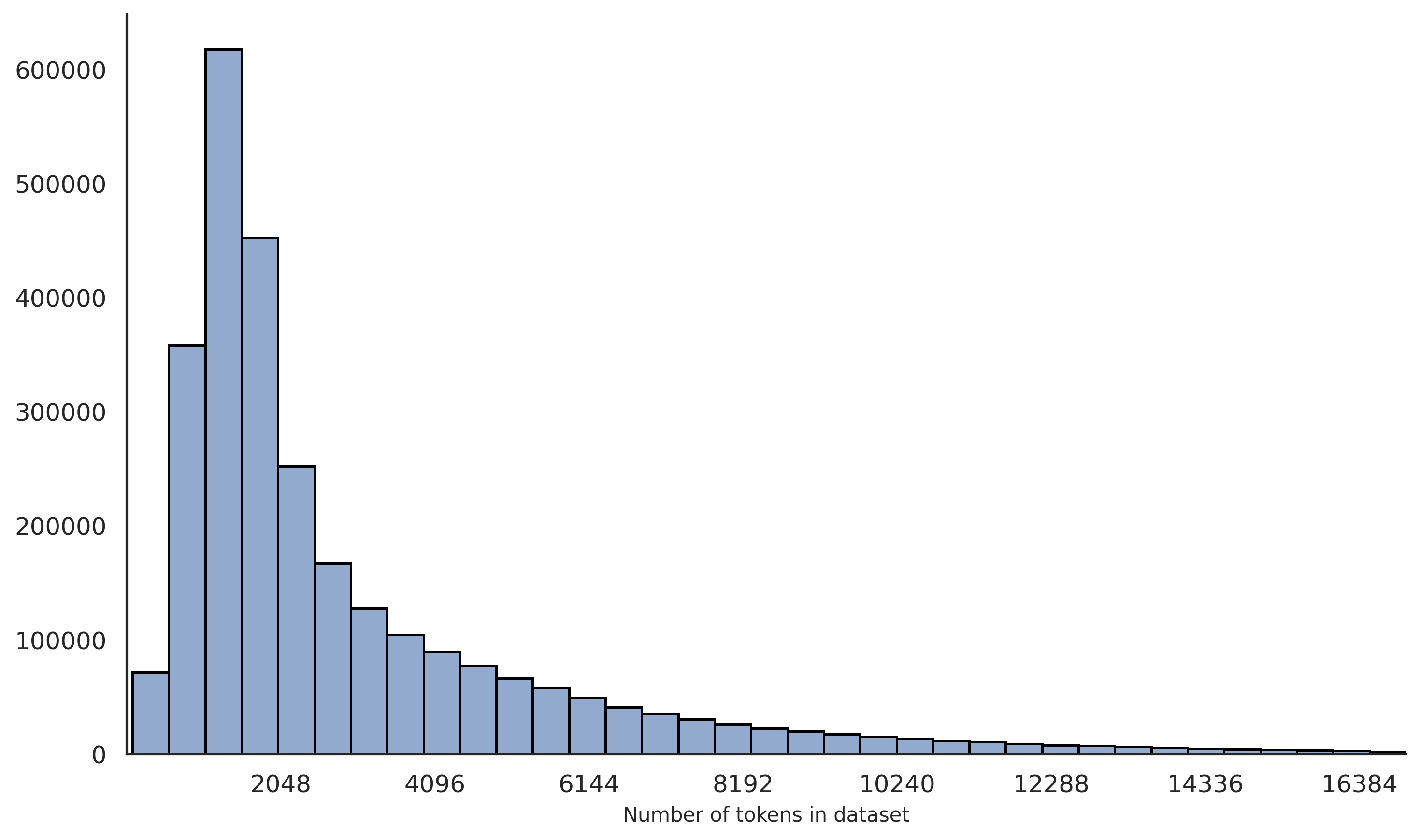}
    \caption{Token length distribution of data entries in the dataset. Most data entries contain fewer than 4096 tokens, with the highest concentration around approximately 2048 tokens. The distribution gradually decreases as the token count increases, indicating fewer samples with longer contexts.}
    \label{fig:enter-label2}
\end{figure}
\subsection{Reference Distribution}
\label{a2}
\begin{figure}[h!]
    \centering
    \includegraphics[width=0.5\linewidth]{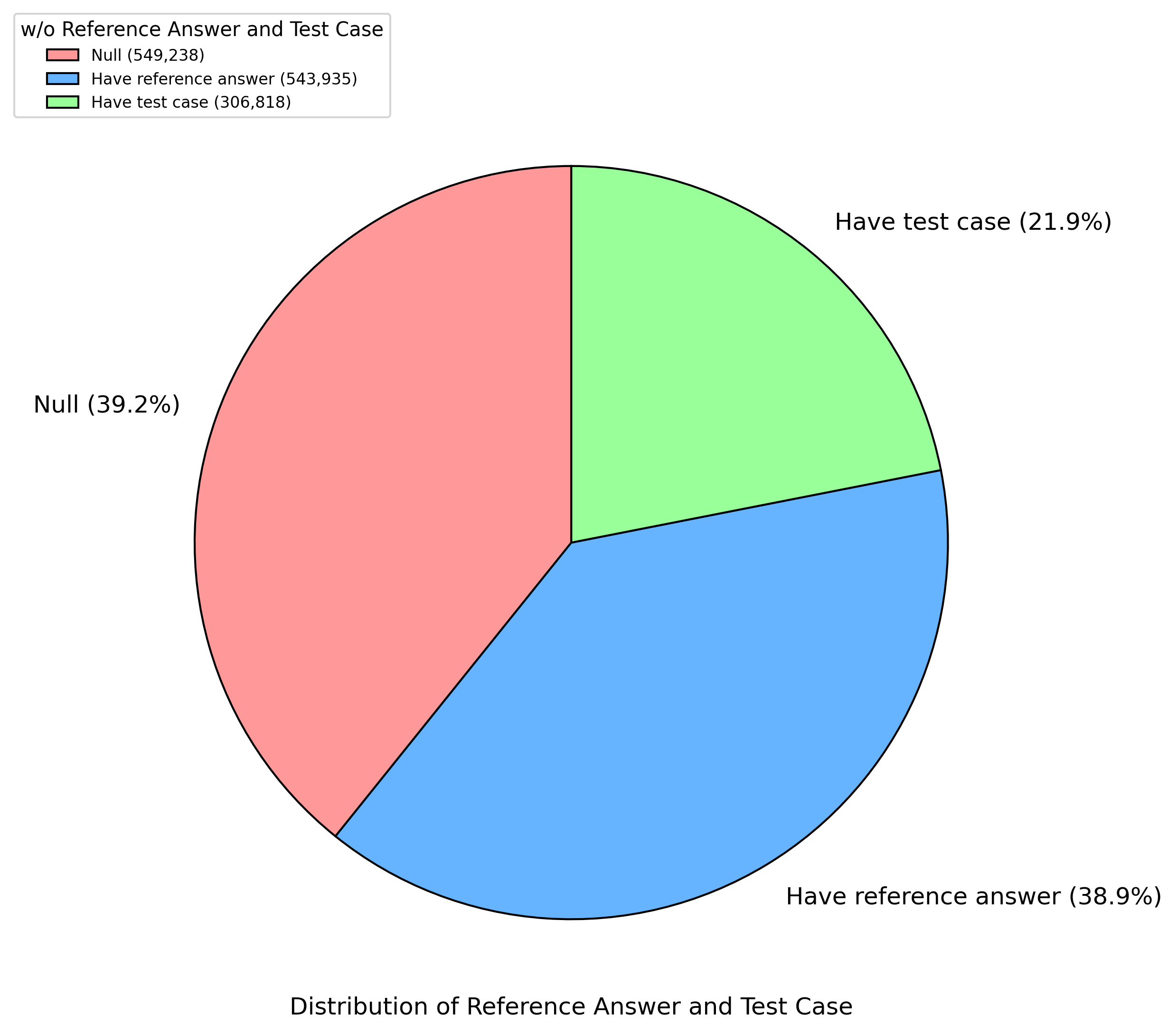}
    \caption{Distribution of reference answers and test cases in the dataset. Among the entries, 38.9\% have reference answers, 21.9\% include test cases, and 39.2\% have neither reference answers nor test cases.}
    \label{fig:enter-label3}
\end{figure}
\subsection{Category Distribution}
\label{a3}
\begin{figure}[H]
    \centering
    \includegraphics[width=0.5\linewidth]{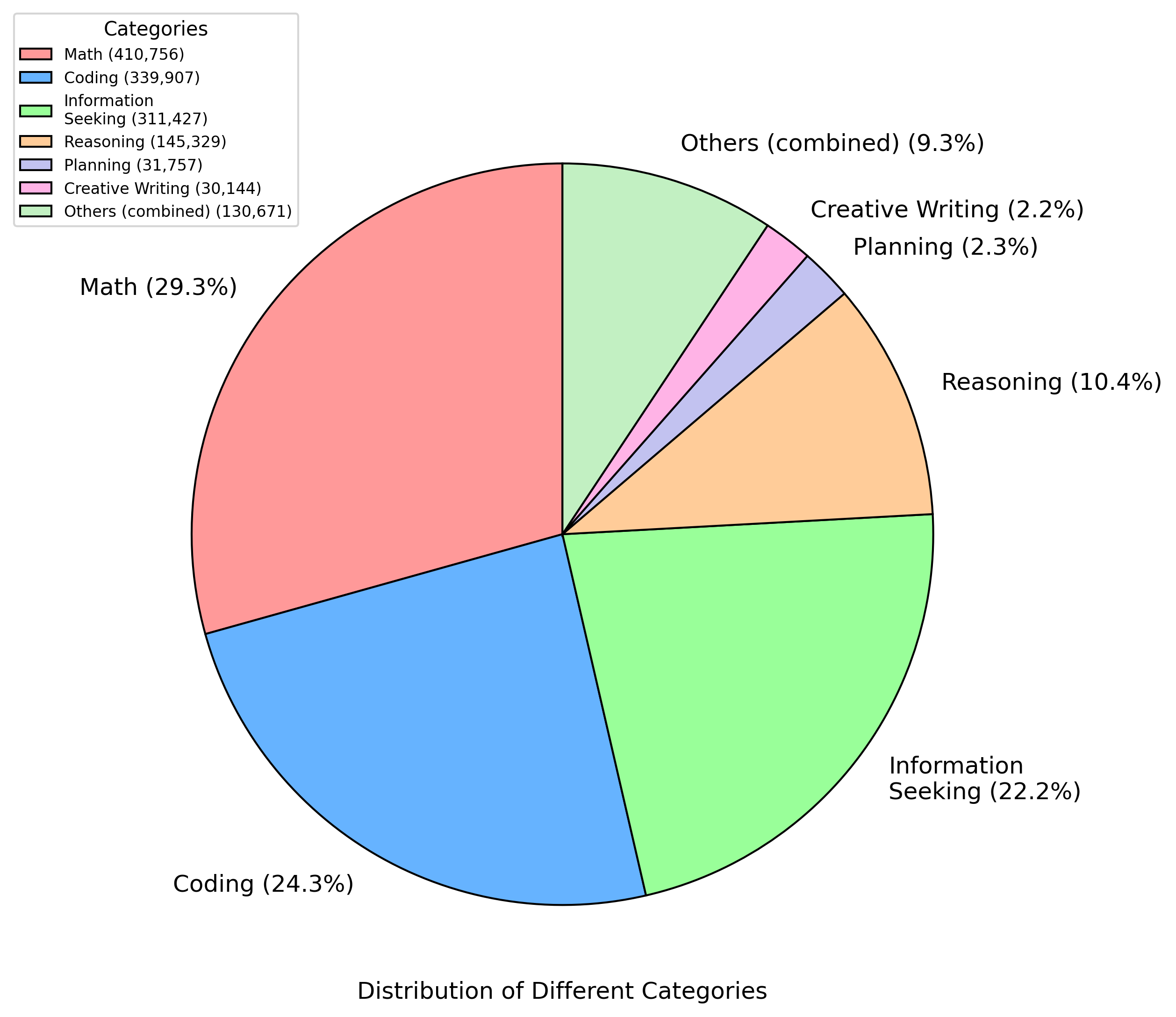}
    \caption{Distribution of data entries across different task categories. The dataset primarily consists of Math (29.3\%), Coding (24.3\%), and Information Seeking (22.2\%) tasks, followed by Reasoning (10.4\%), Planning (2.3\%), Creative Writing (2.2\%), and other combined categories (9.3\%).}
\end{figure}
\subsection{Difficulty Distribution}
\label{a4}
\begin{figure}[h!]
    \centering
    \includegraphics[width=0.5\linewidth]{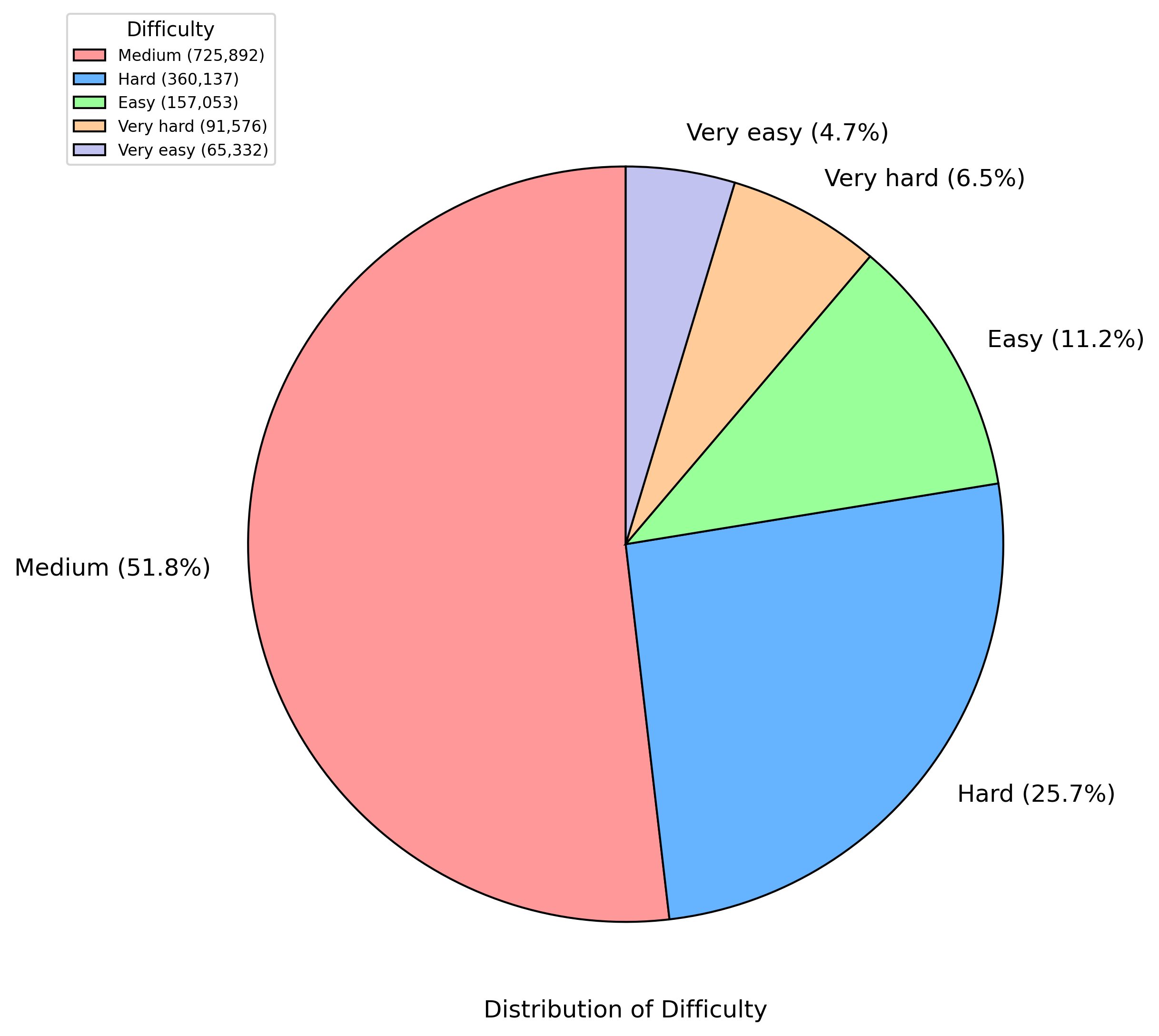}
    \caption{Difficulty distribution of the data entries. Most of the dataset entries are classified as Medium (51.8\%) or Hard (25.7\%). A smaller proportion falls into the Easy (11.2\%), Very Hard (6.5\%), and Very Easy (4.7\%) categories.}
    \label{fig:enter-label4}
\end{figure}

\section{Prompt}

\subsection{Difficulty Rating}
\label{b1}
To grading difficulty rating, wo use prompt as Table~\ref{tab:difficulty rating}.

\begin{table}[!htbp]
  \centering
  \renewcommand{\arraystretch}{0.9} % 缩小行高，让内容更紧凑
  \begin{tabular}{p{\textwidth}}
    \toprule % 专业表格顶线
\small
\begin{lstlisting}
# Instruction

You first need to analyze the given user intent and then label the difficulty level of the user query based on the content of the user query.

## User Query
``` {input} ```

## Evaluation Criteria
Given the user query, you first need to analyze the user intent and the knowledge needed to solve the task in the user query.
Then, rate the difficulty level of the user query as `very easy`, `easy`, `medium`, `hard`, or `very hard`.

Classify the difficulty of the query into one of five levels:
  - very easy: Basic, straightforward questions requiring minimal reasoning.
  - easy: Simple factual queries with slightly more depth.
  - medium: Requires moderate reasoning, explanation, or multi-step processing.
  - hard: Involves advanced concepts, deeper analysis, or multiple interrelated steps.
  - very hard: Expert-level queries demanding significant domain expertise, synthesis, or novel problem-solving.

## Output Format
Just output the json format answer, don't provide additional explanation
Now, please output the difficulty level below in a json format by filling in the placeholders in []:
```json
{
    "difficulty": "[very easy/easy/medium/hard/very hard]"
}
```
\end{lstlisting}
    \\
    \bottomrule % 专业表格底线
  \end{tabular}
  \caption{Difficulty rating prompt.}
  \label{tab:difficulty rating}
\end{table}

\subsection{Category Classification}
\label{b2}
To label category, wo use prompt as Table~\ref{tab:Category labeling}.

\begin{table}[H]
  \centering
  \renewcommand{\arraystretch}{0.9} % 缩小行高，让内容更紧凑
  \begin{tabular}{p{\textwidth}}
    \toprule % 专业表格顶线
\small
\begin{lstlisting}
# Instruction

Please label the task tags for the user query.

## User Query
``` {input} ```

## Tagging the user input
Please label the task tags for the user query. You will need to analyze the user query and select the most relevant task tag from the list below.

all_task_tags = [
    "Logic",  # Queries involving logical puzzles, riddles, or formal deductive reasoning.
    "Information",  # Users ask for specific information or facts about various topics.
    "Editing",  # Involves editing, rephrasing, proofreading, or other tasks related to the composition of general written content.
    "Coding",  # Users seek help with writing, reviewing, or fixing code in programming.
    "Math",  # Queries related to mathematical concepts, problems, and calculations.
    "Brainstorming",  # Involves generating ideas, creative thinking, exploring possibilities, or assisting with decision-making processes.
    "Others"  # Any queries that do not fit into the above categories or are of a miscellaneous nature.
]

## Output Format:
Note that you can only select a single primary tag. Other applicable tags can be added to the list of other tags.
Now, please output your tags below in a json format by filling in the placeholders in <...>:
``` {{
    "primary_tag": "<primary tag>",
    "other_tags": ["<tag 1>", "<tag 2>", ... ]
}}
```
\end{lstlisting}
    \\
    \bottomrule % 专业表格底线
  \end{tabular}
  \caption{Category classification prompt.}
  \label{tab:Category labeling}
\end{table}

\subsection{Correctness Rating}
\label{b3}
To rate correctness, wo use prompt as Table~\ref{tab:correctness rating}.

\begin{table}[H]
  \centering
  \renewcommand{\arraystretch}{0.9} % 缩小行高，让内容更紧凑
  \begin{tabular}{p{\textwidth}}
    \toprule % 专业表格顶线
\small
\begin{lstlisting}
# Instruction
You are an evaluation expert tasked with assessing the correctness of answers provided by a relatively small-sized Language Model (such as a 7B model) based on three inputs. Assign a score from 1 to 5 according to the following criteria:

- Score 5: Completely correct, fully matches the reference answer or accurately addresses the query when the reference answer is not provided.
- Score 4: Mostly correct, minor deviations or insignificant errors that do not affect overall meaning.
- Score 3: Partially correct, includes key information but contains noticeable errors or omissions.
- Score 2: Minimally correct, significant errors or major omissions, answer barely meets requirements.
- Score 1: Completely incorrect, fails to address the question or content severely mismatches the query.

### Please score based on the following inputs:

- **Query**:
  {input_query}

- **Reference Answer:** (May be empty)
  {reference_answer}

- **LLM Answer**:
  {llm_answer}

### Provide your score strictly following the output format below:
```
{{
    "correctness": "<correctness score>",
}}
```
**Justification** (briefly explain your scoring decision):
\end{lstlisting}
    \\
    \bottomrule % 专业表格底线
  \end{tabular}
  \caption{Correctness Rating prompt.}
  \label{tab:correctness rating}
\end{table}

\end{document}